\documentclass[letterpaper]{article} 
\usepackage{colorai25}  
\usepackage{times}  
\usepackage{helvet}  
\usepackage{courier}  
\usepackage[hyphens]{url}  
\usepackage{graphicx} 
\usepackage{natbib}  
\usepackage{caption} 
\usepackage{algorithm}
\usepackage{algorithmic}

\usepackage{newfloat}
\usepackage{listings}
\DeclareCaptionStyle{ruled}{labelfont=normalfont,labelsep=colon,strut=off} 
\floatstyle{ruled}
\newfloat{listing}{tb}{lst}{}
\floatname{listing}{Listing}
\usepackage{amsfonts}
\usepackage{booktabs}
\usepackage{algorithm}
\usepackage{algorithmic}
\usepackage{multirow}
\usepackage{color}

\usepackage{bm}

\usepackage{newfloat}
\usepackage{listings}
\DeclareCaptionStyle{ruled}{labelfont=normalfont,labelsep=colon,strut=off} 
\floatstyle{ruled}
\newfloat{listing}{tb}{lst}{}
\floatname{listing}{Listing}
\title{FinLoRA: Finetuning Quantized Financial Large Language Models \\Using Low-Rank Adaptation on GPUs}
\author{
    Dannong Wang$^1$,  Daniel Kim$^1$,  Bo Jin$^1$,  Xingjian Zhao$^1$,  Tianfan Fu$^1$, \\Steve Yang$^2$, and Xiao-Yang Liu Yanglet$^{1,3}$
}
\affiliations{
    \textsuperscript{\rm 1}Rensselaer Polytechnic Institute, Troy, NY 12180, USA\\
    \textsuperscript{\rm 2}Stevens Institute of Technology, Hoboken, New Jersey, NJ 07030, USA\\
    \textsuperscript{\rm 3}Columbia University, New York, NY 10027, USA\\

}

\begin{document}

\maketitle


\begin{abstract}
Finetuned large language models (LLMs) have shown remarkable performance in financial tasks, such as sentiment analysis and information retrieval. Due to privacy concerns, finetuning and deploying financial LLMs (FinLLMs) locally are crucial for institutions and individuals.  In this paper, we employ quantized low-rank adaptation (QLoRA) to finetune FinLLMs, which leverage low-rank structure and quantization technique to significantly reduce computational requirements while maintaining model performance. We also employ data and pipeline parallelism to enable local finetuning on commodity GPUs. Experiments on financial datasets validate the efficacy of our approach in yielding notable improvements over the base models.

\end{abstract}

\section{Introduction}
Large language models (LLMs) have demonstrated exceptional capabilities in various applications, such as finance~\cite{liu2023data,liu2024differentially}, healthcare~\cite{wang2024twin,lu2024uncertainty,chen2024trialbench}, scientific discovery~\cite{lu2022cot,chen2021data,chen2024uncertainty,fu2024ddn3}, etc. Finetuning using low-rank structures of these models to domain-specific datasets further enhances their performance and improves their applicability to specialized tasks. In the financial domain, finetuned LLMs demonstrate substantial potential for tasks such as sentiment analysis, named entity recognition (NER), and knowledge extraction from financial documents.

FinGPT \cite{liu2023data,liu2024differentially,icdcs,tian2024customized} applied low-rank adaptation techniques for finetuning quantized LLMs in financial contexts, which displayed noticeable improvement over the base model, while having substantial memory reduction and training speedup.  XBRL agent \cite{xbrlagent} evaluated the potential of LLM's capabilities in analyzing XBRL reports. The use of Retrieval-Augmented Generation (RAG) and tools-calling techniques on XBRL-related tasks and demonstrated significant improvement in task accuracy.

Due to sensitive data and regulatory constraints, finetuning and inference of LLMs within local environments remain critical requirements for financial institutions. Furthermore, the ability to create personalized and customized LLMs, finetuned for specific tasks, is essential for maximizing the utility of these models in financial applications, such as the FinGPT search agents \cite{tian2024customized}.




Building upon prior research \cite{icdcs}, we demonstrate that state-of-the-art LLMs can be finetuned for diverse financial tasks locally and cost-effectively using widely accessible GPUs, achieving notable improvements over baseline models.  Our main contributions can be summarized as follows: 


\begin{itemize}
\item We employ Quantized Low-Rank Adaptation (QLoRA) \cite{qlora} to alleviate GPU memory requirements and achieve more efficient finetuning. Using the low-rank structure reduces the number of trainable parameters required for finetuning, and quantization compresses the model size, further reducing GPU memory consumption. 

\item We employ distributed data parallelism (DDP) and pipeline parallelism to leverage multiple GPUs effectively. DDP distributes training data across GPUs to accelerate finetuning, while pipeline parallelism partitions the model at the layer level to optimize memory usage during inference. Together, these strategies enable more efficient fintuning and inference for FinLLMs.

\item We conduct extensive experiments on diverse financial datasets. Models finetuned with QLoRA exhibit up to a 48\% average increase in accuracy compared to baseline models,
which validates the effectiveness of low-rank adaptation and quantization in addressing the unique challenges of FinLLMs. 
\end{itemize}
Our codes are available at Github\footnote{FinLoRA: \url{ https://github.com/YangletLiu/FinLoRA}}



\section{Finetuning LLMs with Quantized Low-rank Adaptation (QLoRA)}


\subsection{Quantized Low-rank Adaptation}

Low-rank adaptation (LoRA) \cite{lora} is a parameter-efficient finetuning method that incorporates a smaller set of trainable weights, such that $\bm{W} = \bm{W}_0 + \Delta \bm{W}$. Let $\bm{W}_0 \in \mathbb{R}^{n \times n}$ denote the pretrained weights and $\Delta \bm{W} = \bm{B}\bm{A}$ denote the update weights, where $\bm{A} \in \mathbb{R}^{r \times n}$ and $\bm{B} \in \mathbb{R}^{n \times r}$ are trainable parameters. Note that $n$ can be large, e.g., $4,096$ and the rank $r \ll n$, say $r = 4, 8$, or $16$. As an example, setting $n = 4,096$, and $r=8$, then $\bm{W}_0$ has approximately $16$ million parameters, while  $\bm{A}$ and $\bm{B}$ together have $65,536$ parameters, which is approximately only 0.039\% the size of $\bm{W}_0$.

During the fine-tuning stage, the forward pass can be expressed as:
\[ \bm{y} = \bm{W}_0 \bm{x} + \Delta \bm{W} \bm{x} = \bm{W}_0 \bm{x} + \bm{B}\bm{A}\bm{x}, \]
where $\bm{W}_0$ denotes the pre-trained weights.

During the inference stage, we do not add $\bm{A}$ and $\bm{B}$ back to $\bm{W}_0$, and we perform
\[ \bm{y} = \bm{W}_0 \bm{x} + \bm{B} \bm{A} \bm{x}.\]
This is different from \cite{lora}, because we will explore the Mixture of Experts approach that trains multiple LoRA adapters. Therefore, it introduces a small amount of additional costs to the inference process.

Quantized LoRA (QLoRA) \cite{qlora} further reduces memory usage by using 8-bit or 4-bit quantization. During finetuning, all weights of the pre-trained model are quantized to 8-bit or 4-bit. Weights will be dynamically dequantized back to 16 bit when performing computation with the input sequence $\bm{x}$ and the adaptor matrix $\bm{A}$ and $\bm{B}$, which remain in $16$-bit precision throughout the process. 

Table 1 illustrates GPU memory usage with QLoRA during finetuning with batch size and rank of 8 and inference with a batch size of 1. The reductions in GPU memory with quantization displayed practical benefits of resource-efficient finetuning and inference for large-scale models.





\newcommand{\blue}{\textcolor{blue}} 
\newcommand{\black}{\textcolor{black}}

\begin{table}[t]
\centering
\caption{GPU memory usage during finetuning (\blue{blue}, batch size and rank of 8) and inference (\black{black}, batch size of 1).}
\begin{tabular}{l|cc}
\toprule
& \multicolumn{2}{c}{GPU memory (GB)} \\
\cmidrule(lr){2-3}
Quantization & Llama 3.1-8B & Llama 3.1-70B \\
\midrule
16-bit & \blue{30.9}, \ \black{15.0} & \blue{$>$300},\ \black{131.5} \\
8-bit & \blue{11.8}, \ \black{8.6}\phantom{0} & \blue{258.2},\ \black{68.5}\phantom{0} \\
4-bit & \phantom{1}\blue{8.7}, \ \black{5.6}\phantom{0} & \phantom{0}\blue{42.8},\ \black{37.8}\phantom{0} \\
\bottomrule
\end{tabular}
\end{table}

\section{High-Performance Optimizations on GPUs}

\subsection{Optimizing Finetuning Process}

To accelerate the finetuning process and leverage the computational power of multiple GPUs, we employed Distributed Data Parallel (DDP), which distributes the training data across GPUs. DDP launches one process per GPU, where each process gets its own copy of the model and optimizer. Each process then receives different inputs, and the gradients are computed and synchronized across all GPUs to accelerate training. DDP provides a substantial speedup when multiple GPUs are available \cite{dpp}.

We also opted to use Brain Floating Point (BF16) during finetuning. BF16 offers the same range of values as FP32 an`d easy conversion to/from FP32. Studies showed that BF16 can achieve similar results as FP32 while having significant speedup and memory savings~\cite{bf16}. 

We used 0/1 Adam optimizer~\cite{01adam}, a modified version of the Adam optimizer that linearize each Adam step and allows utilizing 1-bit compression for faster convergence speed, while offering reduced data volume and higher training throughput. 

\subsection{Optimizing Inference Process}

Inference on large-scale models such as Llama 3.1 70B demands considerable GPU memory resources, particularly when using higher precision like 8-bit or 16-bit. We employ pipeline parallelism, where the model is partitioned at the layer level and distributed across multiple GPUs; each GPU process computes different micro-batches with different parts of the model concurrently \cite{icdcs}.

\section{Experiments}

\subsection{Experimental Setup}

The experiments were conducted on a server equipped with four 16-core AMD EPYC 7313 CPUs, 1 TB of RAM, and four NVIDIA RTX A6000 GPUs, each featuring 48 GB of dedicated GPU memory.

We choose \textbf{Llama 3.1-8B Instruct} and \textbf{Llama 3.1-70B Instruct} \cite{llama} models as base models.

\subsection{Financial Applications}

For general financial tasks, our study focuses on three financial language processing tasks: Sentiment Analysis (SA), Named Entity Recognition (NER), and news headline classification. 
\begin{enumerate}
\item Sentiment Analysis (SA) entails analyzing financial text, such as news articles or tweets, to assign sentiment labels (e.g., positive, negative, or neutral). 
\item Named Entity Recognition (NER) is designed to identify and classify critical entities within financial texts, including organizations, locations, and individuals. 
\item News headline classification involves categorizing headlines according to predefined criteria or questions, facilitating the automated organization and analysis of financial news. 
\end{enumerate}

\noindent For eXtensible Business Reporting Language (XBRL) \cite{xbrl} tasks we focus on tagging and extraction. XBRL is a standardized format designed for the exchange of financial information. Although XBRL documents are based on structured XML (eXtensible Markup Language), their inherent complexity presents challenges that can be addressed using the capabilities of LLMs, thereby facilitating financial reporting and analysis \cite{xbrlagent}.


\subsection{Datasets}

\begin{table}[t]
\centering
\caption{Datasets we used for finetuning and evaluation.}
\begin{tabular}{lcc} 
\toprule
Dataset Name & Type & Train/Test Samples \\
\midrule
FPB & Sentiment Analysis & 1.2K / 3.6K \\
FiQA SA  & Sentiment Analysis & 961 / 150 \\
TFNS & Sentiment Analysis & 9.5K / 2.4K \\
NWGI & Sentiment Analysis & 16.2K / 4.1K \\
Headline & Headline Analysis & 82.2K / 20.5K \\
NER & NER & 13.5K / 3.5K \\

\\[-2.2ex]\hline\\[-1.9ex]
FiNER & XBRL Tagging & 900K / 100K \\
Tags & XBRL Extraction & 300 / 150 \\
Values & XBRL Extraction & 1K / 150 \\
\bottomrule
\end{tabular}
\end{table}

\newcommand{\acc}{\textcolor{blue}} 
\newcommand{\fscore}{\textcolor{black}} 

\begin{table*}[t]
\centering
\caption{Performance on Classification and XBRL Extraction Tasks: Accuracy (\acc{blue}) and F1 Score (\fscore{black}).}
\begin{tabular}{lccccccccccc}
\toprule
\multirow{3}{*}{Model} & \multicolumn{6}{c}{Classification Datasets} & \multicolumn{2}{c}{XBRL Extraction} & \multicolumn{2}{c}{\shortstack{XBRL \\ Tagging}} \\
\cmidrule(lr){2-7} \cmidrule(lr){8-9} \cmidrule(lr){10-11}
& \multirow{2}{*}{FPB} & \multirow{2}{*}{FIQA} & \multirow{2}{*}{TFNS} & \multirow{2}{*}{NWGI} & \multirow{2}{*}{NER} & \multirow{2}{*}{Headline} & \multirow{2}{*}{Tags} & \multirow{2}{*}{Values} & \multirow{2}{*}{FiNER} \\
\\
\midrule
\multirow{2}{*}{Llama-3.1-8B (base)} & \acc{68.73\%} & \acc{46.55\%} & \acc{69.97\%} & \acc{46.58\%} & \acc{48.89\%} & \acc{45.34\%} & \acc{79.37\%} & \acc{55.26\%} & \acc{2.85\%} \\
& \fscore{0.6768} & \fscore{0.5571} & \fscore{0.6834} & \fscore{0.4117} & \fscore{0.5686} & \fscore{0.5576} & \fscore{-} & \fscore{-} & \fscore{-} \\
\addlinespace
\multirow{2}{*}{Llama-3.1-70B (base)} & \acc{74.50\%} & \acc{47.27\%} & \acc{68.42\%} & \acc{79.93\%} & \acc{46.28\%} & \acc{71.68\%} & \acc{89.02\%} & \acc{87.66\%} & \acc{9.41\%} \\
& \fscore{0.7363} & \fscore{0.5645} & \fscore{0.6864} & \fscore{0.7993} & \fscore{0.4539} & \fscore{0.7294} & \fscore{-} & \fscore{-} & \fscore{-} \\
\addlinespace
\multirow{2}{*}{Llama-3.1-8B-4bits-r4} & \acc{\textbf{86.30\%}} & \acc{73.09\%} & \acc{\textbf{88.27\%}} & \acc{80.95\%} & \acc{96.63\%} & \acc{88.03\%} & \acc{\textbf{95.00\%}} & \acc{96.05\%} & \acc{70.45\%} \\
& \fscore{\textbf{0.8600}} & \fscore{0.7811} & \fscore{\textbf{0.8824}} & \fscore{0.8029} & \fscore{0.9664} & \fscore{0.8864} & \fscore{-} & \fscore{-} & \fscore{-} \\
\addlinespace
\multirow{2}{*}{Llama-3.1-8B-8bits-r8} & \acc{82.84\%} & \acc{\textbf{80.36\%}} & \acc{84.05\%} & \acc{\textbf{83.96\%}} & \acc{98.05\%} & \acc{84.66\%} & \acc{94.37\%} & \acc{\textbf{97.36\%}} & \textbf{\acc{75.21\%}} \\
& \fscore{0.8302} & \fscore{\textbf{0.8177}} & \fscore{0.8436} & \fscore{\textbf{0.8492}} & \fscore{0.9806} & \fscore{0.8520} & \fscore{-} & \fscore{-} & \fscore{-} \\
\addlinespace
\multirow{2}{*}{Llama-3.1-70B-4bits-r4} & \acc{80.94\%} & \acc{60.00\%} & \acc{76.01\%} & \acc{80.77\%} & \acc{\textbf{98.88\%}} & \acc{\textbf{96.38\%}} & \acc{-} & \acc{-} & \acc{-} \\
& \fscore{  0.8019} & \fscore{0.6719} & \fscore{0.7219} & \fscore{0.8101} & \fscore{\textbf{0.9887}} & \fscore{\textbf{0.9474}} & \fscore{-} & \fscore{-} & \fscore{-} \\
\bottomrule
\end{tabular}
\end{table*}

\begin{table*}[h!]
\centering
\caption{Finetuning and inference performance on one classification task (NER).}

\begin{tabular}{lcccccc}
\toprule
\multirow{2}{*}{Model} & \multicolumn{4}{c}{Finetuning} & \multicolumn{1}{c}{Inference} \\
\cmidrule(lr){2-5} \cmidrule(lr){6-6} 
 & Batch size & GPU memory (GB) & GPU hours & Adapter size (MB) & Time (s) \\
\midrule
Llama-3.1-8B-4bits-r4  & 16 $\times$ 4  & 83.6 &  0.77 $\times$ 4  & 4.5 & 0.1 \\
Llama-3.1-8B-8bits-r8  & 16 $\times$ 4  & 96.7 & 0.90 $\times$ 4 & 9.0  & 0.1 \\
Llama-3.1-70B-4bits-r4 & 4 $\times$ 4  & 184.3 & 3.50 $\times$ 4 & 21.3  & 0.9 \\
\bottomrule
\end{tabular}

\end{table*}

\begin{table*}[h!]
\centering
\caption{Finetuning and inference performance on XBRL extraction.}

\begin{tabular}{lcccccc}
\toprule
\multirow{2}{*}{Model} & \multicolumn{4}{c}{Finetuning} & \multicolumn{1}{c}{Inference} \\
\cmidrule(lr){2-5} \cmidrule(lr){6-6} 
 & Batch size & GPU memory (GB) & GPU hours & Adapter size (MB) & Time (s) \\
\midrule
Llama-3.1-8B-4bits-r4  & 2 $\times$ 4  & 139.2 & 0.44 $\times$ 4 & 4.5 & 1.9 \\
Llama-3.1-8B-8bits-r8  & 2 $\times$ 4 & 152.2 & 0.48 $\times$ 4 & 9.0  & 1.9 \\
\bottomrule
\end{tabular}

\end{table*}


\subsubsection{Sentiment Analysis}

\begin{itemize}

    \item \textbf{Financial phrasebank (FPB)} \cite{fpb} contains sentences extracted from financial news and reports. These sentences are annotated with sentiment labels. We manually created the train/test split.
    \item \textbf{Financial question-answering sentiment analysis (FiQA SA)} \cite{fiqa} is another sentiment analysis dataset with the same labels as FPB from microblog headlines and financial news.
    \item \textbf{Twitter financial news sentiment (TFNS)} \cite{tfns} comprises annotated tweets related to financial news  labeled with sentiment categories.
    \item \textbf{News with GPT instruction (NWGI)} \cite{liu2023data} comprises samples with seven labels ranging from “strong negative” to “strong positive”. We map the seven labels back to three for simplicity and consistancy with other SA dataset.
\end{itemize}

\subsubsection{Headline classification}
The Headline dataset \cite{headline} categorizes headlines into two classes, "yes" and "no", based on predefined questions.

\subsubsection{Named entity recognition (NER)}
The NER dataset \cite{ner} annotates one entity per sentence, categorized into one of three classes: "location", "person", and "organization"
\subsubsection{XBRL tagging}
The FiNER dataset \cite{loukas-etal-2022-finer} includes sentences annotated with 139 types of XBRL Tags. We processed the dataset so each question comprises of the sentence and one highlighted entity and the answer includes the correct tag.
\subsubsection{XBRL extraction}
The XBRL extraction dataset comprises questions and answers derived from XBRL filings from 2019 to 2023 for Dow Jones 30 companies. Each example includes a question, a text segment from an XBRL file containing the answer, and the ground truth generated using an XBRL file extraction library. From this dataset, we selected the following two tasks:
\begin{itemize}
    \item \textbf{XBRL tag extraction}: The extraction of a specific XBRL tag from a large XBRL raw text segment given a natural language description of the tag.
    \item \textbf{XBRL value extraction}: The extraction of a numeric value from a large XBRL raw text segment given a natural language description of the value. 
    
\end{itemize} 

\noindent To allow better instruction following for the base model, we use one-shot prompting by providing an example question and answer.


\subsection{Implementation Details}
\subsubsection{Finetuning}

We employed distinct finetuning strategies based on the nature of the tasks:

\begin{itemize}
    \item \textbf{General financial tasks and XBRL tagging:} For sentiment analysis, headline classification, and named entity recognition, and XBRL Tagging, single-task fine-tuning was employed.

    \item \textbf{XBRL Extraction:} For XBRL tag extraction and value extraction, multi-task fine-tuning was adopted.
\end{itemize}


\noindent All fine-tuning experiments utilized the 0/1 Adam optimizer \cite{01adam} with learning rate of 1e-4, LoRA alpha of 32, LoRA dropout of 0.1. We use both LoRA rank 4 with 4-bit quantization and rank 8 with 8-bit quantization for Llama 3.1 8B. 
We adjusted the batch size and number of training epochs based on the model size and task:

\begin{itemize}
    \item \textbf{General financial tasks and XBRL tagging:}
    \begin{itemize}
        \item Llama 3.1 8B: Batch size of 16 with gradient accumulation step of 1; 4 epochs.
        \item Llama 3.1 70B: Batch size of 4 with gradient accumulation step of 4; 4 epochs.
    \end{itemize}
    \item \textbf{XBRL extraction:}
    \begin{itemize}
        \item Llama 3.1 8B: Batch size of 2 with gradient accumulation step of 2; 1 epoch.
    \end{itemize}
\end{itemize}


    
        
\subsubsection{Inference}
We use 8-bit quantized inference for all evaluations to ensure consistency. 

\subsection{Performance Metrics}

We evaluate performance using the following metrics:

\subsubsection{Accuracy}

Accuracy is the ratio of the number of correct answers to the total number of queries. An answer is considered correct if the ground truth answer is included in the generated response.

\subsubsection{Weighted F1 score}

For classification tasks, we report the weighted F1 score, calculated as the weighted average of the F1 scores for each class, with weights proportional to the number of instances in each class

\subsubsection{Finetuning and Inference Performance}
\begin{itemize}
    \item \textbf{Batch size}: The batch size per GPU during finetuning. 
    \item \textbf{GPU memory usage}: The sum of the amount of GPU memory used for all GPUs during training. 
    \item \textbf{GPU hours}: The product of total training time and number of GPUs used.
    \item \textbf{Adapter size}: The size of the LoRA adapter file. 
    \item \textbf{Inference speed}: The number of seconds to process an example.
\end{itemize}


\subsection{Results and Analysis}

Tables 3 summarize the accuracy and weighted F1 scores under different finetuning configurations. Table 4 and 5 displays resource usage and inference performance for NER and XBRL extraction. The finetuned Llama 3.1 8B demonstrates noticeable improvements in accuracy compared to its base model and even surpasses the results of the Llama 3.1 70B base model. 

Notably, even with lower quantization (4-bit) and rank 4, the finetuned Llama 3.1 8B model achieves comparable performance to its 8-bit, rank 8 counterpart, while requiring less memory.  Furthermore, the fine-tuned 70B model demonstrates practical usability with 4-bit quantization, showcasing the feasibility of deploying larger LLMs for complex financial tasks in resource-constrained environments.

While we utilized four GPUs to expedite finetuning, it is important to note that all finetuning are achievable with one GPU with 48GB memory, albeit with longer training times.


\section{Conclusion and Future Work}

This study demonstrated the effectiveness of Quantized LoRA (QLoRA) for finetuning large language models (LLMs) for many financial tasks, including sentiment analysis, named entity recognition, news headline analysis, and XBRL filings. We finetuned both Llama 3.1 8B and 70B models on commodity GPUs, achieving up to 48\% improvements in accuracy compared to the base models on average across all tasks. Notably, these performance gains can be achieved with only four GPUs and less than 20 hours of training times per task, making local finetuning and deployment of customized models a feasible option for institutions and individuals.

In future work, we plan to explore multi-task finetuning in classification tasks and expand our investigation of XBRL-related tasks. This will enable FinLoRA to perform more complex analysis and reasoning tasks, further increasing their utility in the financial domain.

\section*{Acknowledgement}

Dannong Wang, Daniel Kim, Bo Jin, Xingjian Zhao, Steve Yang and Xiao-Yang Liu Yanglet acknowledge the support from a NSF IUCRC CRAFT center research grant (CRAFT Grant 22017) for this research. The opinions expressed in this publication do not necessarily represent the views of NSF IUCRC CRAFT. 
Xiao-Yang Liu Yanglet acknowledges the support from Columbia's SIRS and STAR Program, as well as The Tang Family Fund for Research Innovations in FinTech, Engineering, and Business Operations.

\bibliography{aaai25}

\section{Appendix}
\begin{table*}[]
\centering
\caption{Finetuning LLMs with QLoRA methods: listing the number of parameters and GPU memory.}
\begin{tabular}{l|ccc|cc}
\toprule

Model   & Parameters  & \shortstack{GPU Memory \\ Batch = 4} &  \shortstack{GPU Memory \\ Batch = 8} & Model size & Percentage\\
\midrule
Llama3-8B-16bit (base) & 8.03 B  & - & - & 16.06 GB  & - \\

\\[-2.2ex]\hline\\[-1.9ex]

Llama3-8B-r8-16bit & 4.72 M  & 30.91 GB & 30.91 GB& 16.08 GB & 100.1\%\\
Llama3-8B-r8-8bit & 4.72 M  & 11.41 GB&  11.81 GB& 8.04 GB & 50.1\% \\
Llama3-8B-r8-4bit & 4.72 M  & 8.26 GB& 8.65 GB& 4.02 GB & 25.0\% \\

\\[-2.2ex]\hline\\[-1.9ex]

Llama3-8B-r4-16bit & 2.36 M  & 30.90 GB& 30.90 GB& 16.07 GB & 100.1\% \\
Llama3-8B-r4-8bit & 2.36 M  & 11.40 GB& 11.78 GB& 8.03 GB & 50.0\%\\
Llama3-8B-r4-4bit & 2.36 M  & 8.25 GB& 8.61 GB& 4.02 GB  & 25.0\% \\

\\[-2.0ex]\hline\hline\\[-1.6ex]

Llama3-70B-16bit (base) & 70.56 B & - & - & 151.53 GB & - \\

\\[-2.2ex]\hline\\[-1.9ex]
Llama3-70B-r8-16bit & 22.28 M  & - & - & 151.57 GB & 100.0\% \\
Llama3-70B-r8-8bit & 22.28 M  & 173.57 GB & 258.17 GB  & 75.79 GB & 50.0\% \\
Llama3-70B-r8-4bit & 22.28 M  & 42.78 GB& 42.78 GB& 37.89 GB & 25.0\% \\

\\[-2.2ex]\hline\\[-1.9ex]

Llama3-70B-r4-16bit & 11.14 M  & - & - & 151.55 GB & 100.0\% \\
Llama3-70B-r4-8bit & 11.14 M  & 173.36 GB & 258.11 GB  & 75.70 GB & 50.0\% \\
Llama3-70B-r4-4bit & 11.14 M &  42.73 GB& 42.73 GB &  37.89 GB & 25.0\%    \\

\bottomrule
\end{tabular}
\end{table*}

\end{document}